%% file: index.tex
\documentclass[11pt,a4paper]{article}
\usepackage[hyperref]{acl2019}
\usepackage{times}
\usepackage{latexsym}
\usepackage{url}
\usepackage{import}
\usepackage{multirow}
\usepackage{multicol}
\usepackage{makecell}
\usepackage{booktabs}
\usepackage{xcolor}
\usepackage{colortbl}
\usepackage{subcaption}
\usepackage{amsmath}

\usepackage{graphicx}
\usepackage{enumitem}
\aclfinalcopy 
\def\aclpaperid{2502} 
\title{Exploring Author Context for Detecting Intended vs Perceived Sarcasm}
\author{Silviu Vlad Oprea \\
  School of Informatics\\
  University of Edinburgh \\
  Edinburgh, United Kingdom\\
  \texttt{silviu.oprea@ed.ac.uk} \\\And
  Walid Magdy \\
  School of Informatics\\
  University of Edinburgh \\
  Edinburgh, United Kingdom\\
  \texttt{wmagdy@inf.ed.ac.uk} \\}
\date{}
\newcommand{\q}[1]{\bgroup\color{red}[QUESTION: #1]\egroup}
\newcommand{\walid}[1]{\bgroup\color{blue}[Walid: #1]\egroup}
\newcommand{\bonnie}[1]{\bgroup\color{blue}[Bonnie: #1]\egroup}
\newcommand{\todo}[1]{\bgroup\color{blue}[TODO: #1]\egroup}
\newcommand{\changed}[1]{\bgroup\color{red}#1\egroup}
\newcommand{\place}{\bgroup\color{green!70!black}[PLACEHOLDER]\ \egroup}

\begin{document}
\maketitle
%
%
\begin{abstract}
\label{sec:abstract}
\import{sections/}{abstract.tex}
\end{abstract}
%
%
\section{Introduction}
\import{sections/}{introduction.tex}
%
%
\section{Background}
\label{section:related-work}
\import{sections/}{related-work.tex}
%
%
\section{Sarcasm Datasets}
\label{section:data}
\import{sections/}{data.tex}
%
%
\section{Contextual Sarcasm Detection Models}
\label{section:models}
\import{sections/}{models.tex}
%
%
\section{Effect of Context on Sarcasm Detection}
\label{section:experiments}
\import{sections/}{experiments.tex}
%
%
\section{Conclusion}
\label{section:conclusion}
\import{sections/}{conclusion.tex}
%
%
\section{Acknowledgements}
\label{section:acknowledgements}
\import{sections/}{acknowledgements.tex}
%
%

\fontsize{9.1pt}{10.1pt}
\selectfont 
\bibliography{index}
\bibliographystyle{aaai}
\end{document}

%% file: sections/abstract.tex
We investigate the impact of using author context on textual sarcasm detection. We define author context as the embedded representation of their historical posts on Twitter and suggest neural models that extract these representations. We experiment with two tweet datasets, one labelled manually for sarcasm, and the other via tag-based distant supervision. We achieve state-of-the-art performance on the second dataset, but not on the one labelled manually, indicating a difference between intended sarcasm, captured by distant supervision, and perceived sarcasm, captured by manual labelling.

%% file: sections/introduction.tex
Sarcasm is a form of irony that occurs when there is a discrepancy between the literal meaning of an utterance and its intended meaning. This discrepancy is used to express a form of dissociative attitude towards a previous proposition, often in the form of contempt or derogation~\citep{WILSON20061722}.

Sarcasm is omnipresent on the social web and can be highly disruptive of systems that harness this data~\citep{L14-1527}. It is therefore imperative to devise model for textual sarcasm detection. The effectiveness of such models depends on the quality of labelled data used for training. Two methods are commonly used to label texts for sarcasm: manual labelling by human annotators; and tag-based distant supervision. In the latter, texts are considered sarcastic if they contain specific tags, such as \#sarcasm and \#sarcastic.

Most work on computational sarcasm detection extracts lexical and pragmatic cues available in the text being classified~\citep{campbell2012there,r3,riloffembeddings,tay-att}. However, sarcasm is a contextual phenomenon and detecting it often requires prior information about the author, audience and previous interactions between them, that originates beyond the text itself~\citep{Context2}.

In this work we investigate the impact of author context on the current sarcastic behaviour of the author. We identify author context with the embedded representation of their historical tweets. We use the term \emph{user} to refer to the author of a tweet and the phrase \emph{user embedding} to refer to such a representation. Given a tweet $t$ posted by user $u^t$ with user embedding $e^t$, we address two questions: (1) Is $e^t$ predictive of the sarcastic nature of $t$? (2) Is the predictive power of $e^t$ on the sarcastic nature of $t$ the same if $t$ is labelled via manual labelling vs distant supervision?

To our knowledge, previous research that considers author context~\citep{rajadesignan-2015,firstcontextualised,amir,cascade} only experiments on distant supervision datasets.
We experiment on datasets representative of both labelling methods, namely Riloff~\citep{r3}, labelled manually, and Ptacek~\citep{ptacek}, labelled via distant supervision.

We suggest neural models to build user embeddings and achieve state-of-the-art results on Ptacek, but not on Riloff. Comparing and analyzing the discrepancy, our findings indicate a difference between the sarcasm that is \emph{intended} by the author, captured by distant supervision, represented in Ptacek, and sarcasm that is \emph{perceived} by the audience, captured by manual labelling, represented in Riloff. This difference has been highlighted by linguistic and psycholinguistic studies in the past~\citep{intended-perceived-linguistics-2,char-THIS}, being attributed to socio-cultural differences between the author and the audience. However, up to our knowledge, it has not been considered in the context of sarcasm detection so far. Our work suggests a future research direction in sarcasm detection where the two types of sarcasm are treated as separate phenomena and socio-cultural differences are taken into account.

%% file: sections/related-work.tex
\subsection{Sarcasm Detection}
Based on the information considered when classifying a text as sarcastic or non-sarcastic, we identify two classes of models across literature: local models and contextual models.
\paragraph{Local Models}
Local models only consider information available within the text being classified. Most work in this direction considers linguistic incongruity~\citep{campbell2012there} to be a marker of sarcasm. \citet{r3} consider a positive verb used in a negative sentiment context to indicate sarcasm.
\citet{riloffembeddings} use the cosine similarity between embedded representations of words. Recent work attempts to capture incongruity using a neural network with an intra-attention mechanism~\citep{tay-att}.
\paragraph{Contextual Models}
Contextual models utilize both local and contextual information.
There is a limited amount of work in this direction.
\citet{wallace-etal-2015-sparse}, working with Reddit data, include information about the forum type where the post to be classified was posted.
For Twitter data, \citet{rajadesignan-2015} and \citet{firstcontextualised} represent user context by a set of manually-curated features extracted from their historical tweets.
\citet{amir} merge all historical tweets of a user into one historical document and use the Paragraph Vector model~\citep{doc2vec} to build a representation of that document.
Building on their work, \citet{cascade} extract in addition personality features from the historical document.
Despite reporting encouraging results, these models are only tested on datasets labelled via distant supervision. In our work, we compare the performance of our models when tested on datasets representative of both manual annotation and distant supervision.
\subsection{Intended vs Perceived Sarcasm}
\citet{Context3} notice a lack of consistence in how sarcasm is defined by people of different socio-cultural backgrounds.  As a result, an utterance that is intended as sarcastic by its author might not be perceived as such by audiences of different backgrounds~\citep{Context2}. When a tweet is sarcastic from the perspective of its author, we call the resulting phenomenon \emph{intended sarcasm}. When it is sarcastic from the perspective of an audience member, we call the phenomenon \emph{perceived sarcasm}.

%% file: sections/data.tex
We test our models on two popular tweet datasets, one labelled manually and the other via distant supervision.
\subsection{Riloff dataset}
The Riloff dataset consists of 3,200 tweet IDs.
These tweets were manually labeled by third party annotators. The labels capture the subjective perception of the annotators (perceived sarcasm). Three separate labels were collected for each tweet and the dominant one was chosen as the final label.

We attempted to collect the corresponding tweets using the Twitter API\footnote{https://developer.twitter.com}, as well as the historical timeline tweets for each user, to be used later for building user embeddings. For a user with tweet $t$ in Riloff, we collected those historical tweets posted before $t$. Only 701 original tweets, along with the corresponding user timelines, could be retrieved. Others have either been removed from Twitter, the corresponding user accounts have been disabled, or the API did not retrieve any historical tweets.

Table~\ref{table:all-labels} shows the label distribution across this dataset. We divided the dataset into ten buckets, using eight for training, one for validation and one for testing. The division into buckets is stratified by users, i.e. all tweets from a user end up in the same bucket. Stratification makes sure any specific embedding is only used during training, during validation, or during testing.
We further ensured the overall class balance is represented in all of the three sets.
Table~\ref{table:all-labels} shows the size of each set.
\begin{table*}[t]
    \centering
    \small
    \begin{tabular}{@{}l | c c c | c c c@{}} 
       \toprule
       dataset & size & sarcastic & non-sarcastic & train & valid & test\\
       \midrule
       Riloff & 701 & 192 & 509 & 551 & 88 & 62 \\
       Ptacek & 27,177 & 15,164 & 12,013 & 21,670 & 2,711 & 2,797 \\ 
       \bottomrule
    \end{tabular}
    \caption{Label distribution across our datasets; and distribution into train, validation and test sets.}
    \label{table:all-labels}
\end{table*}
\subsection{Ptacek dataset}
The Ptacek dataset consists of 50,000 tweet IDs labelled via distant supervision. Tags used as markers of sarcasm are \#sarcasm, \#sarcastic, \#satire and \#irony. This dataset reflects intended sarcasm, since the original poster tagged their own tweet as sarcastic through the hashtag.

In a similar setting as with Riloff we could only collect 27,177 tweets and corresponding timelines.
%
We divided them into ten buckets and stratified by users. During preprocessing we removed all sarcasm-marking tags from both the training tweets and the historical tweets. Table~\ref{table:all-labels} shows statistics on both datasets.

%% file: sections/models.tex
\begin{figure*}
    \centering
    \includegraphics[width=0.6\textwidth]{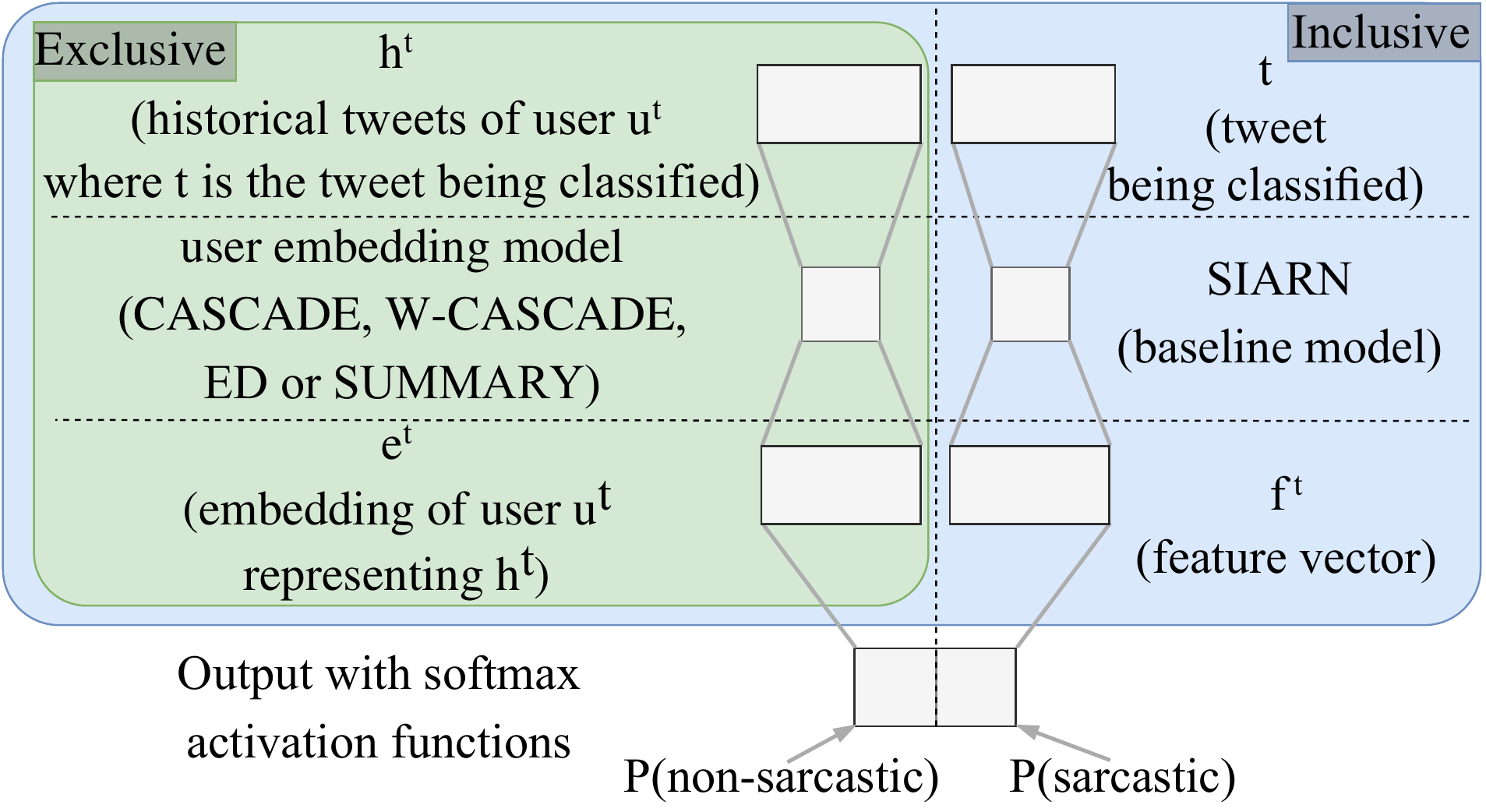}
    \caption{The architecture of the models used. Exclusive models do not use the current tweet being classified, prediction being based solely on user history. Inclusive models use both user history and the current tweet.}
    \label{figure:models}
\end{figure*}
%
Let $T$ be a set of tweets. For any $t\in T$, let $u^t$ be the user who posted tweet $t$. Let $h^t$ be a set of historical tweets of user $u^t$, posted before $t$, with $h^t\cap T=\emptyset$ and let $e^t$ be the embedding of user $u^t$, i.e. a vector representation of $h^t$.
Let $Y=\{$sarcastic, non-sarcastic$\}$ be the output space.
Our goal is to find a model $m:\{(t, e^t)|t\in T\}\rightarrow Y$.

As a baseline, we implement the SIARN (Single-Dimension Intra-Attention Network) model proposed by \citep{tay-att}, since it achieves the best published results on both our datasets. SIARN only looks at the tweet being classified, that is $\textrm{SIARN}(t, e^t)=m'(t)$.

Further, we introduce two classes of models: exclusive and inclusive models.
In exclusive models, the decision whether $t\in T$ is sarcastic or not is independent of $t$, i.e. $m(t, e^t)=m'(e^t)$. The content of the tweet being classified is not considered, prediction being based solely on user historical tweets. The architecture of such a model is shown in Figure~\ref{figure:models}.
We feed the user embedding $e^t$ to a layer with softmax activations to output a probability distribution over $Y$. We name these models EX-[emb], where [emb] is the name of the user embedding model.

Inclusive models account for both $t$ and $e^t$, as shown in Figure~\ref{figure:models}. We start with the feature vector $f^t$ extracted by SIARN from $t$. 
We then concatenate $f^t$ with $e^t$ and use an output layer with softmax activations. We name these models IN-[emb], where [emb] is the user embedding model.
We now look at several user embedding models that build $e^t$ for a user $u^t$ as a representation of $h^t$.
Recall that $\forall u\in usr(T): hist(u)\cap T=\emptyset$, where $usr(T)$ is the image of $T$ under $usr$.
\paragraph{CASCADE Embeddings}
Up to our knowledge, the user embedding model that has proven most informative in a sarcasm detection pipeline so far is CASCADE~\citep{cascade}. However, it has only been tested on a dataset of Reddit\footnote{https://www.reddit.com} posts labelled via distant supervision. We test it on our datasets. Following original authors, we merge all tweets from $h^t$ in a single document $d^t$, giving corpus $C=\{d^t|t\in T\}$. Using the Paragraph Vector model~\citep{doc2vec} we generate a representation $v^t$ of $d^t$. Next, we feed $d^t$ to a neural network pre-trained on the personality detection corpus released by~\citet{MATTHEWS1999583}, which contains labels for the Big-Five personality traits~\citep{goldberg1993structure}.
We merge the resulting hidden state $p^t$ of the network with $v^t$ using Generalized Canonical Correlation Analysis (GCCA) as described by~\citet{cascade} to get $e^t$.
\paragraph{W-CASCADE Embeddings}
CASCADE treats all historical tweets in the same manner. However, as studies in cognitive psychology argue \citep{ltwm}, long-term working memory plays an important role in verbal reasoning and textual comprehension. We therefore expect recent historical tweets to have a greater influence on the current behaviour of a user, compared to older ones.
To account for this, we suggest the following model that accounts for the temporal arrangement of historical tweets. We first use CASCADE to build $v^t_r$ and $p^t_r$, and to merge them into $e^t_r$ using GCCA, $\forall r\in h^t$.
We then divide the sequence $\langle e^t_{r_1},e^t_{r_2},\ldots, e^t_{r_{|h^t|}}\rangle$ into ten contiguous partitions and multiply each vector with the index of the partition it belongs to. That is, we multiply $e^t_{r_i}$ by $i\ \%\ |h^t|+1$, where $\%$ is the modulus operator. By convention, the tweet with the highest index is the most recent one. Finally, we sum the resulting vectors and normalize the result to get $e^t$.
\paragraph{ED Embeddings}
One of the main advantages of the encoder-decoder model~\citep{ed}, commonly used for sequence prediction tasks, is its ability to handle inputs and outputs of variable length. The encoder, a recurrent network, transforms an input sequence into an internal representation of fixed dimension. The decoder, another recurrent network, generates an output sequence using this representation.
We use bi-directional LSTM cells%
~\citep{brnn} and identify $e^t_{r_i}$, $1\le i\le |h^t|$, with the internal state of the encoder after feeding in $r_i$. The training objective is to reconstruct the input $r_i$. We employ the same weighting technique as we did for W-CASCADE to construct $e^t$.
%
\paragraph{SUMMARY Embeddings}
We use an encoder-decoder model as in the previous paragraph, but change the objective from reconstructing the input to summarizing it. We pre-train the model on the Gigaword standard summarization corpus\footnote{\url{https://github.com/harvardnlp/sent-summary}}.

%% file: sections/experiments.tex
%
\subsection{Experimental Setup}
We filter out all tweets shorter than three words and replace all words that only appear once in the entire corpus with an UNK token. Then, we encode each tweet as a sequence of word vectors initialized using GloVe embeddings~\citep{glove2014}.
Following the authors SIARN, our baseline, we set the word embedding dimension to 100.
We tune the dimension of all CASCADE embeddings to 100 on the validation set. For comparability, we set W-CASCADE embeddings to the same dimension.
For CASCADE embeddings we make use of the implementation available at \url{https://github.com/SenticNet/cascade}. 
When training ED and SUMMARY, our decoder implements attention over the input vectors. We use the general global attention mechanism suggested by~\citet{D15-1166}. We implement both ED and SUMMARY using the OpenNMT toolkit~\citep{klein-etal-2017-opennmt}.

For comparability with SIARN, our baseline, we follow its authors in setting a batch size of 16 for the Riloff dataset, and of 512 for the Ptacek dataset, and in training for 30 epochs using the RMSProp optimizer~\citep{rmsprop} with a learning rate of 0.001.
Our code and data can be obtained by contacting us.
\subsection{Results}
\begin{table}[t]
    \centering
    \small
    \begin{tabular}{l l | c c} 
        \toprule
        \multicolumn{2}{c|}{{Model}} & {Riloff} & {Ptacek}\\
        \midrule
        & SIARN (baseline) & 0.711 & 0.863 \\ \midrule
        \multirow{2}{*}{exclusive} & EX-CASCADE & 0.457 & 0.802 \\
        & EX-W-CASCADE & 0.478 & \textbf{0.922}\\
        & EX-ED & \textbf{0.546} & 0.873 \\
        & EX-SUMMARY & 0.492 & 0.845 \\\midrule
        \multirow{2}{*}{inclusive} & IN-CASCADE & 0.723 & 0.873 \\
        & IN-W-CASCADE & 0.714 & \textbf{0.934} \\
        & IN-ED    & \textbf{0.739} & 0.887 \\
        & IN-SUMMARY   & 0.679 & 0.892 \\
        \bottomrule
    \end{tabular}
    \caption{F1 score achieved on the Riloff and Ptacek datasets for both exclusive and inclusive models. 
    Best results for each model class are highlighted in \textbf{bold}.
    }
   \label{table:exin-results}
\end{table}
\begin{table}[t]
    \centering
    \small
    \begin{tabular}{l c c} 
        \toprule
        \multicolumn{1}{c}{Model} & Riloff & \#Riloff\\
        \midrule
        EX-CASCADE & 0.457 & 0.818 \\
        EX-W-CASCADE & 0.478 & 0.797 \\
        EX-ED & \textbf{0.545} & \textbf{0.827} \\
        EX-SUMMARY & 0.492 & 0.772 \\
    \bottomrule
    \end{tabular}
    \caption{F1 score achieved by the exclusive models on the \#Riloff dataset, compared to Riloff dataset. Best results are highlighted in \textbf{bold}.}
    \label{table:tag-results}
    \label{table:results}
\end{table}
All results are reported in Table~\ref{table:exin-results}.
User embeddings show remarkable predictive power on the Ptacek dataset. In particular, using the EX-W-CASCADE model, we get better results (f1-score 0.922) than the baseline (f1-score 0.863) without even looking at the tweet being predicted.
On the Riloff dataset, however, user embeddings seem to be far less informative, with EX-W-CASCADE yielding an f1-score of only 0.478.
Out of the exclusive models, we get the highest f1-score of 0.546 using EX-ED on Riloff. By contrast we get 0.873 on Ptacek using EX-ED.

The state-of-the-art performance of exclusive models on Ptacek indicate that users seem to have a prior disposition to being either sarcastic or non-sarcastic, which can be deduced from historical behaviour. However, this behaviour can change over time, as we achieve better performance when accounting for the temporal arrangement of historical tweets, as we do in W-CASCADE.

On the Riloff dataset the performance of exclusive models is considerably lower. In the following, we investigate the possible reasons for this large difference in performance between the two datasets. 
\subsection{Performance Analysis}
\label{subsection:performance-analysis}
Riloff dataset is annotated manually, which might not reflect the intention of the users, but rather the subjective perception of the annotators. In this light, we could expect user embeddings to have poor predictive power. Perhaps annotator embeddings would shed more light.
\begin{table}[t]
\centering
\small
    \begin{tabular}{lcc}
        \toprule
                                &                         with tag &                  without any tag\\\hline
           labelled sarcastic  & \cellcolor{green!45!gray!45} 190 & \cellcolor{red!45!gray!45}     2\\
        labelled non-sarcastic & \cellcolor{red!45!gray!45}   217 & \cellcolor{green!45!gray!45} 292 \\ 
        \bottomrule
    \end{tabular}
    \caption{
    Disagreement between manual labels and the presence of sarcasm tags in the Riloff dataset, as discussed in Section~\ref{subsection:performance-analysis}.
    }
    \label{table:riloff-disagreement}
\end{table}

We noticed that many of the tweets in Riloff contain one or more of the tags that were used to mark sarcasm in Ptacek. For all tweets in Riloff, we checked the agreement between containing such a tag, and being manually annotated as sarcastic. The results are shown in Table~\ref{table:riloff-disagreement}.
Note that the statistics shown are not for the entire dataset as published by~\citet{r3}, but for the subset of tweets coming from users without blocked profiles and from which we could gather historical tweets, as discussed in Section~\ref{section:data}.
We notice a large disagreement.
In particular, 217 out of the 509 tweets that were annotated manually as non-sarcastic contained such a tag. 
The lack of coherence between the presence of sarcasm tags and manual annotations in the Riloff dataset suggests that the two labelling methods capture distinct phenomena, considering the subjective nature of sarcasm. Previous research in linguistics and psycholinguistics~\citep{intended-perceived-linguistics-2,char-THIS} attributes this difference to socio-cultural differences between the author and the audience and shows that the difference persists even when contextual information is provided.

To investigate further, we re-labelled the Riloff dataset via distant supervision considering these tags as markers of sarcasm, to create the \#Riloff dataset. We applied the exclusive models on \#Riloff and noticed a considerably higher predictive power than on Riloff. Results are reported in Table~\ref{table:tag-results}.
Author history seems therefore predictive of authorial sarcastic intention, but not of external perception. This could indicate that future work should differentiate between the two types of sarcasm: intended and perceived. Both are important to detect, for applications such as opinion mining for the former and hate speech detection for the latter.

%% file: sections/conclusion.tex
We studied the predictive power of user embeddings in textual sarcasm detection across datasets labelled via both manual labelling and distant supervision.
We suggested several neural models to build user embeddings, achieving state-of-the-art results for distant supervision, but not for manual labelling. 
We account for discrepancy by reference to the different type of sarcasm captured by the two labelling methods, attributed by previous research in linguistics and psycholinguistics~\citep{intended-perceived-linguistics-2,char-THIS} to socio-cultural differences between the author and the audience. We suggest a future research direction in sarcasm detection where the two types of sarcasm are treated as separate phenomena and socio-cultural differences are taken into account.

%% file: sections/acknowledgements.tex
This work was supported in part by the EPSRC Centre for Doctoral Training in Data Science, funded by the UK Engineering and Physical Sciences Research Council (grant EP/L016427/1); the University of Edinburgh; and The Financial Times.